\RequirePackage{xcolor}
\documentclass{article}

\usepackage[final]{corl_2022} 

\usepackage{amssymb, bm, mathtools}
\usepackage[pdftex, xetex]{graphicx}
\usepackage{enumerate, setspace}
\usepackage{float, colortbl, tabularx, longtable, multirow, subcaption, environ, wrapfig, textcomp, booktabs}
\usepackage{pgf, tikz, framed}
\usepackage[normalem]{ulem}
\usetikzlibrary{arrows,positioning,automata,shadows,fit,shapes}
\usepackage[english]{babel}
\usepackage{microtype}
\microtypecontext{spacing=nonfrench}
\usepackage{algorithm}
\usepackage{algorithmic}
\usepackage{amsmath}
\usepackage{times}
\usepackage{array}
\usepackage{wrapfig}

\usepackage{soul}
\usepackage{enumitem}
\usepackage{gensymb}
\usepackage{textcomp}
\usepackage{url}

\usepackage[colorinlistoftodos]{todonotes}
\newcommand{\jd}[1]{}

\long\def\ignorethis#1{}

\renewcommand{\eqref}[1]{Eqn~(\ref{eq:#1})}

\newcolumntype{L}{>{\centering\arraybackslash}m{2cm}}

\renewcommand\vec{\mathbf}

\DeclareMathOperator{\E}{\mathbb{E}}

\newcommand{\edc}[1]{{#1}}

\newcommand{\projectwebsite}{https://sites.google.com/view/lirf-corl-2022/}

\title{Training Robots to Evaluate Robots: Example-Based Interactive Reward Functions for Policy Learning}

\author{Kun Huang \hskip2em Edward S. Hu \hskip 2em Dinesh Jayaraman\\
    GRASP Lab, University of Pennsylvania \\\texttt{\{huangkun, hued, dineshj\}@seas.upenn.edu} \\
}

\begin{document}

\maketitle



\begin{abstract}
Physical interactions can often help reveal information that is not readily apparent. For example, we may tug at a table leg to evaluate whether it is built well, or turn a water bottle upside down to check that it is watertight. We propose to train robots to acquire such interactive behaviors automatically, for the purpose of evaluating the result of an attempted robotic skill execution. These evaluations in turn serve as ``interactive reward functions'' (IRFs) for training reinforcement learning policies to perform the target skill, such as screwing the table leg tightly. In addition, even after task policies are fully trained, IRFs can serve as verification mechanisms that improve online task execution. For any given task, our IRFs can be conveniently trained using only examples of successful outcomes, and no further specification is needed to train the task policy thereafter.
In our evaluations on door locking and weighted block stacking in simulation, and screw tightening on a real robot, IRFs enable large performance improvements, even outperforming baselines with access to demonstrations or carefully engineered rewards. {Project website: \url{\projectwebsite}}
\end{abstract}

\vspace{-3mm}
\section{Introduction}\label{sec:intro}
\vspace{-2mm}
Consider a kitchen robot that must perform a large number of tasks, such as opening a refrigerator, cutting vegetables, tightening a water bottle lid, or flipping a pancake. How might this robot acquire these skills? Common skill acquisition approaches involve heavy engineering and expertise for each skill, directed either towards developing model-based control policies, or towards specifying dense reward functions for reinforcement learning (RL). These approaches do not scale well to the goal of acquiring large numbers of skills. Consequently, there is growing interest in scalable approaches for RL-based skill acquisition that permit easy task specification by non-experts.

A particularly promising task specification framework 
that we call exemplar rewards, involves specifying tasks merely by showing the robot learner what the environment should look like after a well-executed skill~\cite{fu2018variational,  xie2018few, singh2019end,eysenbach2021replacing, konyushkova2020semi, zolna2020offline, ma2022smodice}. For example, to specify the task of opening a refrigerator door, it would suffice to show the learner some images of open refrigerators. Then, once the skill is specified, the robot performs RL with the aim of altering the environment to generate observations similar to those examples. 
When this framework is successful, a human user teaching the robot does not need technical expertise and does not need to even demonstrate full behaviors for task execution. Instead, they need only capture images of the outcome. 

\begin{figure}[t]
\centering
\includegraphics[width=0.95\linewidth]{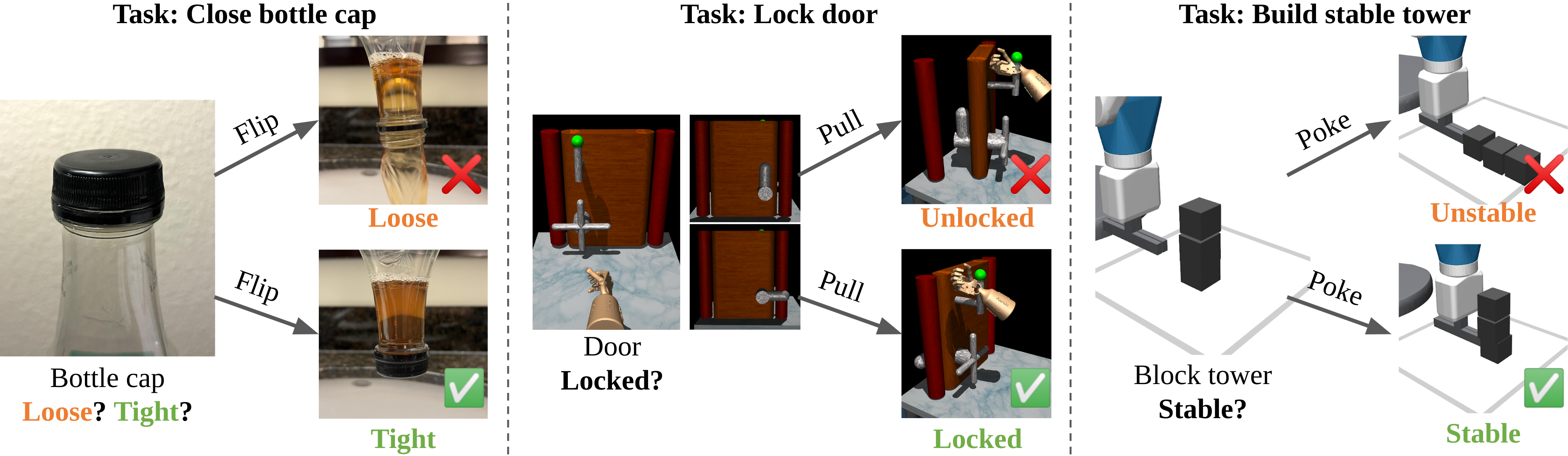}
\caption{In each of these three settings, task success / failure cannot be determined from passive image observations, but is easily revealed through appropriate interactions.
}
\label{fig:concept}
\vspace{-5mm}
\end{figure}

Although this framework is appealing, many tasks cannot be specified by image observations of task success, or any other fixed sensor setup.
For example, consider the task of closing a water bottle, as in Figure \ref{fig:concept}. What image might specify a successful execution of this skill? An image of a water bottle with a slightly loose cap would look identical to one with a tight cap, but a bottle closing policy that does not fully tighten the cap would be useless. Thus, in this setting, an observation of the task success state from a camera constitutes only a partial observation and does not suffice to specify the task. Specifically, when non-success states might look visually similar to success states, image-based task specification and assessment fails. 
Figure \ref{fig:concept} shows a few examples of this state aliasing problem: Is the bottle cap tight? Is the door locked properly? Is the tower of objects with unknown masses stable? There are many more: Is a screw tight? Is the leg of the table secured well? Does the smoothie have the right consistency?
None of these questions can be answered correctly from passive image observations.  



The failures of passive image-based assessment in these settings preclude robotic skill learning from image examples of task success. 
To fix this while retaining the versatility and user-friendliness of the exemplar rewards framework, we propose an \textit{interactive} approach for evaluating task success. As motivation, consider how a human might turn a water bottle upside down to confirm that it does not leak, and is therefore closed well. Here, the interaction reveals the key determinant of task success, namely, whether or not the cap is watertight. 

We design an approach to automatically learn such interaction behaviors to evaluate task success. Our examples are no longer mere images, but instead \textit{real actionable physical instances}.
In the water bottle setting, rather than image snapshots of closed water bottles, our approach receives actual physical bottles with tightly closed caps. The robot can interact with these objects, such as by turning them upside down, to discover what about them constitutes successful execution of the desired ``bottle closing'' skill (i.e., the bottle does not leak) and how they behave differently from the outcomes of unsuccessful execution (i.e., it leaks). In the other examples in Figure \ref{fig:concept}, a robot might try to pull the door open or nudge the tower to evaluate whether the task is complete.

We make two key contributions. First, we propose an approach to train \textit{interactive reward functions}, which are policies that specify what it means to correctly perform a skill and thus drive the improvement of the task policy learners in a reinforcement learning framework. Next, we show how those same reward policies can be reused to enable \textit{in-the-loop introspective verification} behaviors during online skill execution, making robots more scrupulous and reliable. We evaluate these contributions on three simulated and real robotic control settings, and show substantial improvements in task policy learning and execution. 

\vspace{-3mm}
\section{Background and Problem Setup}\label{sec:background}
\vspace{-2mm}
\subsection{Classifier-Based Exemplar Rewards in fully observed MDPs}
\vspace{-2mm}
Reinforcement learning (RL) methods offer the promise of scalable data-driven synthesis of robot controllers for arbitrary new tasks. Consider robotic task settings formalized as Markov decision processes (MDPs), defined by the tuple $(\mathcal{S}, \mathcal{A}, \mathcal{T}, R, \mu)$. At each time $t$, an agent selects an action $a_t \in \mathcal{A}$ in state $s_t \in \mathcal{S}$, transitions to the next state $s_{t+1}$ with probability $\mathcal{T}(s_{t+1} |
s_t, a_t)$, and receives reward $r_t=R(s_t, a_t, s_{t+1})$. Thus, the agent emits actions into the environment and receives two things in return: new state observations and rewards. Ignoring discounting, a good RL task policy $\pi_T (a_t | s_t)$ selects actions to maximize the sum of rewards over time $\sum_t r_t$, starting at a state drawn from an initial state distribution $\mu$.

Thus, it is the reward function $R$ that specifies what task to perform in a given environment, and RL approaches in practice require expertise and effort for reward engineering~\cite{singh2009rewards} to specify each new task, avoiding misspecification and guiding efficient learning. 
This often takes the form of tuned weighted combinations of many carefully constructed heuristic terms in the RL reward objective~\cite{molchanov2019sim, Rudin2022-aq, peng2018deepmimic, gym-github}, or even privileged task-specific sensors added to the environment during training~\cite{schenck2017visual, yahya2017collective}. To circumvent laborious manual reward design, many methods aim to learn rewards from data. Inverse reinforcement learning methods~\cite{ng2000algorithms, ramachandran2007bayesian, arora2021survey, ho2016generative} learn task reward functions from optimal demonstrations, but such demonstrations are typically expensive and may even be impossible to obtain. 
Other methods train RL agents by learning rewards explicitly or implicitly from interactive human feedback~\cite{knox2009interactively, veeriah2016face, chernova2014robot, christiano2017deep, lin2020review}, but these have the drawback of requiring in-the-loop queryable human teachers. 

We build upon the popular ``exemplar rewards'' framework, which explores task specifications purely through examples~\cite{fu2018variational,  xie2018few, singh2019end,eysenbach2021replacing, konyushkova2020semi, zolna2020offline, ma2022smodice}. In classifier-based exemplar rewards approaches~\cite{fu2018variational, singh2019end}, the human teacher trains a classifier $\hat{R}(s)$ to label task outcomes as successes or failures, by gathering some success examples manually, and then generating failure examples from the task policy as it trains. Armed with this learned reward function $\hat{R}(s)$, the agent trains its task policy $\pi_T(a|s)$ to generate success outcomes and thereby maximize the estimated rewards. An appealing property of this framework is that training a policy for each new task only requires a specification of the final goal state that should be achieved, without any need to specify how to achieve it.

Our work extends this exemplar rewards framework: as motivated in Section~\ref{sec:intro}, rather than relying on passive observations, we will train interactive evaluation policies to evaluate task success, and these evaluations will in turn serve as ``interactive reward functions'' for training task policies. This permits applying exemplar rewards to ``partially observed'' settings that do not permit accurate task success evaluation from images alone. We expand on the partial observability problem below.



\vspace{-3mm}
\subsection{Partial Observability, State Aliasing, and History-Based POMDP Policies} \label{sec:POMDP}
\vspace{-2mm}

Why are classifier-based exemplar rewards approaches not suited to settings such as the door locking task in Figure~\ref{fig:concept} (middle)? The root cause is ``partial observability.'' To solve this task, the robot must push the door into the closed position and then rotate the latch into the locking position. It only sees images from a fixed camera in front of the door. The latch, however, is occluded, and the robot can only observe and act upon the four handles on the front of the door, which rotate the latch through an axle mechanism. The handles are visually identical, so any visual configuration of the handles corresponds to any one of four configurations of the latch behind the door.
While this setting is constructed  as a pedagogical example to expose and study the state aliasing problem, it is representative of several real-world settings, as discussed in Section~\ref{sec:intro} and Figure~\ref{fig:concept}. 

Such settings are modeled as partially observable MDPs, or POMDPs~\cite{kaelbling1998planning}. At each time, the agent only partially observes the underlying latent task state $s_t$ (i.e. the latch configuration), through an observation $o_t \sim O (o | s_t)$, where the observation function $O$ generates samples from the observation space $\Omega$ (i.e, images of the handle configuration). The original task state $s_t$ is not directly observable. Thus, the task POMDP is defined by $(\mathcal{S}, \mathcal{A}, \mathcal{T}, R, \Omega, O, \mu)$. 

Several prior works deal with the question of how the task policy in a POMDP might map to good actions despite only seeing partial observations at each time. The key to this is the assumption, very often reasonable, that while individual observations are indeed incomplete, observation \textit{histories} contain much more information about the state. Then, a POMDP policy can perform well by relying on a history of observations and actions, either directly~\cite{mccallum1993overcoming,hausknecht2015deep,zhu2017improving, ni2021recurrent} as $\pi_T(a_t | o_{t-H:t}, a_{t-H-1:t-1})$, or by first estimating a belief distribution $b_t(s_t | o_{t-H:t}, a_{t-H-1:t-1})$ over the state and then mapping to actions~\cite{kaelbling1998planning,gregor2019shaping,gangwani2020learning,weisz2018sample,Igl2018-sm}, as $\pi_T(a_t | b_t)$. In the door locking setting, the robot might observe the latch position before closing the door. Therefore, even if the latch is occluded after the door is closed, a history-based policy could still track its position. In other words, if trained well with the right reward function, such history-based POMDP policies can still learn good task behaviors.


\vspace{-3mm}
\section{Exemplar Interactive Reward Functions}\label{sec:approach}
\vspace{-2mm}
This brings us to the key question: how to specify reward functions $R$ for training POMDP policies? In particular, could we extend the appealing exemplar rewards framework to POMDPs? 

First, note that partial observability has to do with what \textit{policies} can see, not where \textit{rewards} come from. Therefore, it is feasible in theory to provide state-based exemplar rewards $R(s)$ that are not limited by partial observability. This implies relying on additional sensors for reward estimation during training that are not available to the policy $\pi_T$ that is being trained. In other words, while the states $s_t$ are no longer observable to $\pi_T$, it may still be possible to learn exemplar rewards $\hat{R}(s_t)$ through privileged instrumentation of the environment during training to allow the reward function to observe the full state $s_t$. For example, privileged thermal cameras could 
measure fluid levels to train an RGB image-based fluid pouring policy~\cite{schenck2017visual}. However, such additional instrumentation is labor-intensive and task-specific, and sacrifices the appealing scalability and ease of use of the exemplar rewards framework.

How might we do this? Learning exemplar rewards $\hat{R}(o_t)$ based on the agent's instantaneous observations would suffer from state aliasing: in the door locking setting, such a reward function would not be able to distinguish a correctly latched door from one in which the latch is off by a factor of 90\degree! Nor does the exemplar reward learning framework permit relying on observation histories, like the solution we described above for POMDP \textit{policies}. To see this, recall that exemplar rewards are trained from samples of successful task \textit{outcomes} rather than full demonstrations, so there is no observation-action history information available for training $\hat{R}(o_{t-H:t}, a_{t-H-1:t-1})$.

\vspace{-2mm}
\subsection{Actionable Examples and Interactive Reward Functions (IRFs)}

\begin{figure}[t]
\centering
\includegraphics[width=\linewidth]{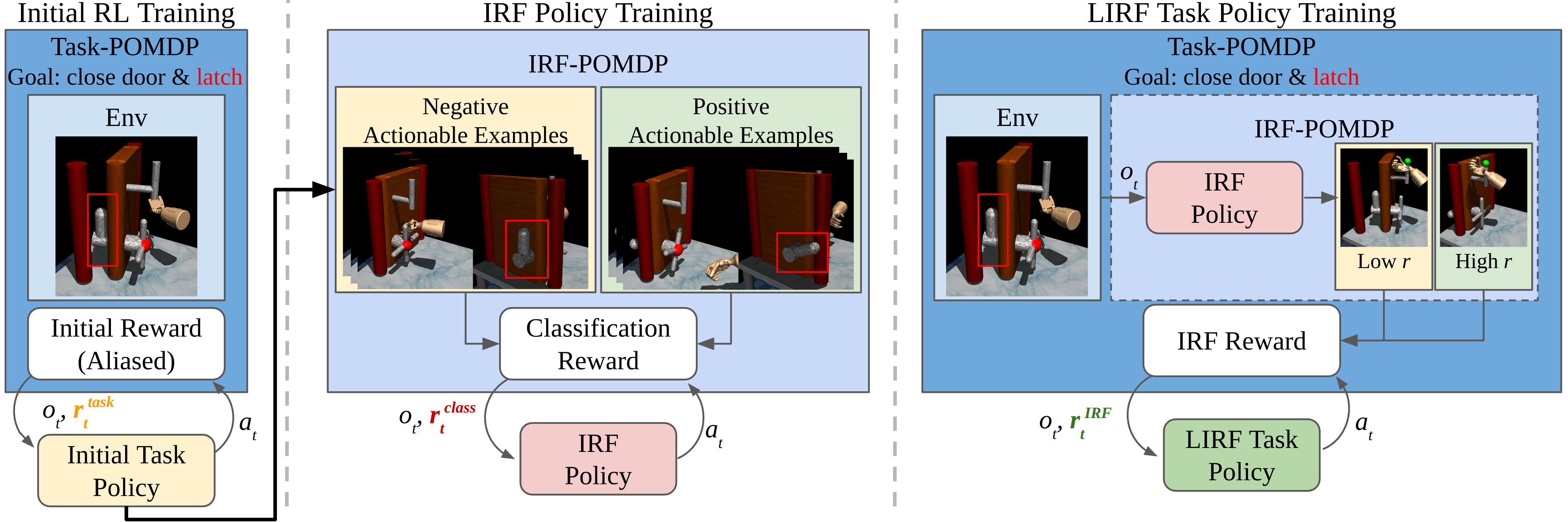}
\caption{\textbf{Method Overview.} In a partially observable environment, (left) we first learn an initial task policy $\pi_T^0$ using passive classifier-based rewards. (middle) Then we train an IRF policy $\pi_R$ to distinguish between provided ``actionable positive examples'' and $\pi_T^0$-generated negative examples. (right) Finally, we use $\pi_R$ to provide the correct rewards for training a LIRF task policy $\pi_T$.
}
\label{fig:method}
\vspace{-3mm}
\end{figure}

Our primary contribution is a solution to this conundrum: rather than presenting successful outcome examples as singular pre-recorded observations, we propose to present them as \textit{``actionable examples''}, with which the robot can interact and generate new observations. In other words, to learn a reward function for locking a door, the robot gets access to several physical instances of successfully locked doors. Then, instead of training passive reward functions from image observations, we propose learning \textit{``interactive reward functions''} (IRF), which consist of robot action policies $\pi_{R}$ that reveal the task rewards. See the schematic in Figure~\ref{fig:method}. We expand upon this idea below.


First, we gather ``actionable success examples'' $P = \{\sigma^+_1, ..., \sigma^+_K\}$ and ``actionable failure examples'' $N = \{\sigma^-_1, ..., \sigma^-_K\}$. $P$ and $N$ contain object or environment configurations: in Figure \ref{fig:concept}, each $\sigma^+$ might be a tight bottle cap, a well-locked door, a stable tower, and so on.  Now, the IRF policy may be seen as the solution to a new ``IRF POMDP''. This is identical to the task POMDP, except for two key differences. First, it is initialized to an actionable example state $s_0 = \sigma \sim P \cup N$. Next, its reward function $R$ is based on discovering the label of $s_0$: was it drawn from $P$ or from $N$? All other elements $(\mathcal{S}, \mathcal{A}, \mathcal{T}, \Omega, O)$ of the task POMDP are retained.\footnote{In some experiments, we use different actions  $\mathcal{A}$ for the IRF policy $\pi_R$ than for the task policy 
$\pi_T$.}

Algorithm~\ref{algo:main_algo} shows pseudocode for our ``Learning from IRFs" (LIRF) approach.
We point out three key features here.
\textbf{(1)} First, we bootstrap the task policy $\pi_T$ by training against a single-observation-based reward classifier $D(o_t)$. This corresponds to running prior exemplar reward approaches; we use VICE~\cite{fu2018variational}.  
\textbf{(2)} Next, in keeping with the exemplar rewards framework, we provide only the positive examples $P$. Negative examples $N$ are instead generated by the above-initialized task policy. Thus, the IRF policy $\pi_R$ serves as a GAN-like adversarial critic~\cite{goodfellow2014generative, ho2016generative} for training the task policy $\pi_T$. 
\textbf{(3)} Finally, how could we discover the true task completion reward, i.e., the label $P$ (corresponding to positive reward) or the label $N$ (negative reward) at the end of IRF execution? For this, we could, in theory, train a classifier from the full IRF trajectory history. Instead, we find that it suffices to reuse the single-observation classifier $D(o_t)$ from (1) above, and apply it only to the final state after executing $\pi_R$. Intuitively, $\pi_R$ learns to modify the environment state $\sigma \xrightarrow{\pi_R} s^*$ such that $o^* \sim O(s^*)$ reveals whether $\sigma$ was drawn from $P$ or $N$.  For example, in door locking, $\pi_R$ might learn to tug at the door: if the door is correctly locked ($P$), it stays closed, and if it is not locked ($N$), it opens. Thus, a single post-IRF observation suffices to classify $\sigma$. Figure~\ref{fig:method} shows a schematic.


\begin{algorithm}[t]
\small
\caption{Learning from Interactive Reward Functions (LIRF) Framework}
\begin{algorithmic}[1]
\renewcommand{\algorithmicrequire}{\textbf{Require:}}
\REQUIRE a set $P$ of positive actionable examples that specify successful task execution.
\STATE Following image-based exemplar rewards approaches such as VICE~\cite{fu2018variational}, train a single-observation reward $D(o) \rightarrow [0, 1]$ and an initial task policy $\pi_T^0$  
in the task POMDP.
\STATE Rollout $\pi_T^0$ for $n$ times, collecting negative actionable outcomes in $N$.
\STATE Train an IRF policy $\pi_R$ in the IRF POMDP, where the environment is initialized from state $\sigma \sim P \cup N$, to maximize the classification reward:
\begin{align}
     \operatorname*{\E}_{s_0 \in P} \bigg[ \sum_t \log (D(o_t)) \bigg] 
     + \operatorname*{\E}_{s_0 \in N} \bigg[ \sum_t \log (1 - D(o_t)) \bigg]
\label{eq:pi_v_rew}
\end{align}

\STATE Train the final LIRF task policy $\pi_T$ in the task POMDP, modified to include the additional terminal state reward
$\hat{R}(o_T) =  D(o_T) + \lambda D(o^*)$,
where $o_T$ is the observation for the terminal state $s_T$, and $o^*$ is generated by the IRF policy $\pi_R$ as $s_T \xrightarrow{\pi_R} s^*$, $o^* \sim O(s^*)$. The final reward function uses the dense single-observation reward for intermediate states, and the terminal state reward for the final state.
\begin{align}
\hat{R}(o_t) =
\begin{cases}
D(o_T) + \lambda D(o^*) & \text{if $t = T$} \\
D(o_t) & \text{otherwise}
\end{cases}
\label{eq:pi_t_rew}
\end{align}


\end{algorithmic}
\label{algo:main_algo}

\end{algorithm}

\vspace{-2mm}
\subsection{IRFs as Verification Mechanisms for Introspective, Fault-Tolerant Behaviors}
Next, we observe that IRFs $\pi_R$ rely on no privileged information beyond the task policy $\pi_T$'s own observations. Thus, while their primary purpose is to provide reward functions for training task policies, IRFs can also be deployed together with the task policies after training. We propose a simple approach to use IRFs as verification mechanisms \textit{in-the-loop} during task execution. Once the task policy $\pi_T$ is executed for a fixed episode length, we will run the IRF $\pi_R$ to determine whether the task is complete. If it is detected as incomplete, we will resume executing the task policy. This perform-verify cycle continues until the IRF is satisfied that the task is complete (or timeout). 
Appendix \ref{appendix:lirf_verify} contains pseudocode for this procedure.



\vspace{-3mm}
\section{Other Related Work}\label{sec:related}
\vspace{-2mm}
LIRF trains task policies through rewards computed by an interactive reward function, which is itself a robot policy trained from examples. While Section~\ref{sec:background} extensively discussed connections to reward design, exemplar rewards, and POMDP policy learning to set up our approach, we now position our contributions against two other key related areas of prior work.

\textbf{Adversarial Robot RL.~~} Like LIRF, some prior works train a target task policy against another ``adversary'' policy~\cite{pinto2017robust, pan2019risk, ma2018improved, pinto2017supervision}. 
\citet{pinto2017supervision} train a grasping robot to select more stable grasps by competing against an adversary robot that tugs at the grasped objects to dislodge them. 
Rather than adversary policies that destabilize the task policy to increase robustness, we train adversary policies that provide outcome-based rewards to train the task policy. 



\textbf{Interactive Perception and Verification.~~} Our method draws on ideas from interactive perception~\cite{bohg2017interactive}, where agents in partially observed settings act to obtain information about the latent state.  Interactive task verification mechanisms (see \cite{kroemer2021review} for a survey) employ human-specified interactive perception behaviors to assess the state of a task. For example, a robot may use engineered motion primitives or perturbations to detect and correct task failures in manipulation~\cite{su2018learning, niekum2015online}. In contrast, we \textit{learn} interactive verification behaviors from examples, and use them not only during task execution, but also to define the reward function for training task skill policies. One way to frame our contribution here is as follows: in reinforcement learning, agents emit actions into the environment and receive in return not just new states, but also \textit{task rewards}. To our knowledge, LIRF is the first approach that employs interactive perception to directly estimate task rewards, rather than the state.




\vspace{-3mm}
\section{Experiments}
\vspace{-2mm}
We design various simulated and real-world experiments to evaluate LIRF for example-based task specification in partially observable tasks that manifest the state aliasing issue.
We aim to answer the following questions: 
(1) Does our interactive reward learning framework enable us to learn policies to solve partially observable tasks that alternative approaches cannot, with comparable or even greater task supervision? (2) Does the learned IRF policy sensibly evaluate task completion and provide good rewards? (3) Can using the IRF policy in-the-loop for task verification improve task execution performance?


\begin{figure}[t]
\centering
\includegraphics[width=\linewidth]{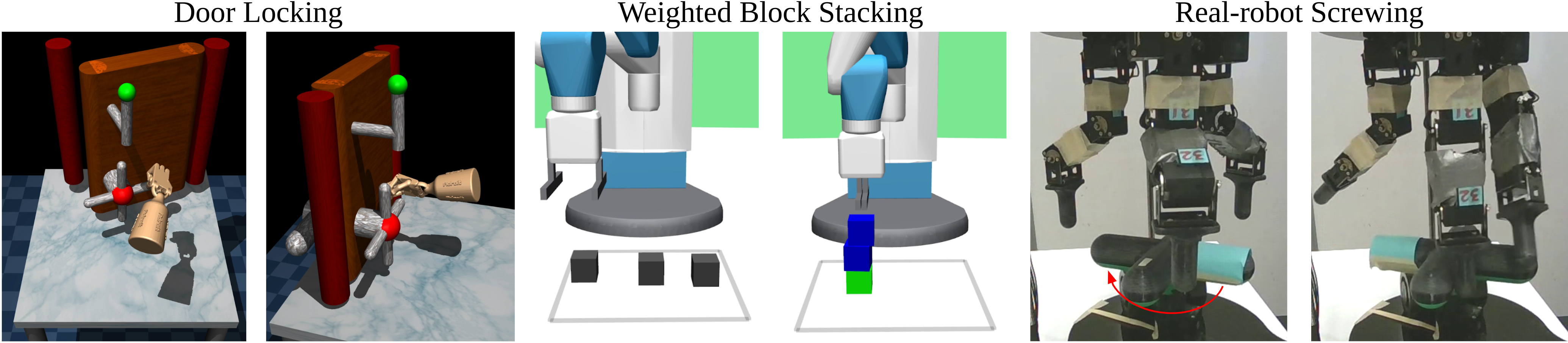}
\caption{Our three tasks.
For each pair of 2 images, we show the task initialization on the left, and the goal state on the right.
In the middle panel, the 3 blocks are visually identical but they differ in weight. The heaviest block is green for visualization. In the right panel, we put a blue mark on the valve for visualizing the orientation.
}
\vspace{-5mm}

\label{fig:envs}
\end{figure}


\paragraph{Task Setups.}
We evaluate LIRF on 3 tasks, illustrated in Figure \ref{fig:envs}, and described below. See Appendix \ref{appendix:environment} for more details and \edc{the project website} for experiment videos. 
\jd{Have we released code for these envs? Kun: the github repo is linked on the website}
\begin{itemize}[leftmargin=*]
\item  \textbf{Door Locking (Sim):~~} 
We instantiate the door locking task from sections \ref{sec:POMDP} and \ref{sec:approach} with an Adroit Hand using MuJoCo \cite{rajeswaran2017learning}. The robot has a fixed viewpoint in front of the door, and must close the door and rotate the symmetric, four-handled doorknob to fully lock it (see Figure \ref{fig:envs}).
The observation space mimics perfect vision: it consists of the door orientation, the door frame position, the \textit{aliased} cross-handle orientation [$0$, $90\degree$)\footnote{The cross-handle is symmetrical and looks visually identical at 90\degree offsets, e.g. 10\degree, 100 \degree, 190 \degree, 280 \degree} and the Adroit hand pose.
The door is initialized as open, and the position of the door frame and the orientation of the latch are randomized.

\item \textbf{Weighted Block Stacking (Sim):~~} 
We further test our algorithm on a weighted block stacking task based on the MuJoCo Fetch environment~\cite{brockman2016openai} as shown in Figure \ref{fig:envs}.
The goal of this task is to stack 3 visually identical blocks into a ``stable'' tower. One block significantly heavier than the other two blocks, so the optimal strategy is to place it at the bottom. 
Again, the observation space mimics perfect vision: only the unordered poses of blocks are observable. The weight of a block is only revealed after it is picked up, to mimic physically plausible ``hefting'' behavior.

\item \textbf{Screw Tightening (Real):~~} 
To examine the robustness and generality of our algorithm, we test it on a real-robot screwing experiment using a D'Claw \cite{ahn2020robel} that has 9 joints (Figure \ref{fig:envs}). The objective of this task is to turn the 4-prong valve clock-wise for around $180\degree{}$ into the ``tightened" state (white line on the valve base).
We engage a motor underneath the valve to mimic screw locking. 
Again, the observation consists of the historical aliased valve angles, i.e. $[0, 90\degree{})$ and robot joint angles. 






\end{itemize}

\textbf{LIRF Implementation Details.~~}
We use the soft actor-critic (SAC) algorithm \cite{haarnoja2018soft} to train both the task policies and the IRF policies for each experiment and report results with mean and variance over 5 seeds for each simulated task.
The weight $\lambda$ for the sparse IRF reward in Eq~\ref{eq:pi_v_rew} needs to be ``large enough'' (Appendix \ref{sec:lambda}) to have a substantial effect on LIRF training, but beyond this, the algorithm is not very sensitive to $\lambda$; we set $\lambda=1000$. 
All policies are trained until convergence, which takes around $2$M simulation steps for door locking and block stacking LIRF policies, $50$k steps for screwing LIRF policies ($\sim10$ hours on the real robot), and around $20$k steps for all IRF policies. 
More details in Appendix \ref{appendix:lirf_training}. We will release all code and environments.

\paragraph{Baselines.}
For evaluating our LIRF algorithm, we compare against RL policies learned with VICE exemplar rewards~\cite{fu2018variational}
and with ground truth state-based rewards (``GT State Reward'').
In addition, we compare against GAIfO \cite{torabi2018generative}, an imitation-from-observation algorithm that learns from 
demonstrations. 
We provide as many demonstrations to GAIfO as the number of episodes with positive examples provided to LIRF (namely, 250, 10k, and 100 for door locking, weighted block stacking, and screwing respectively). Note that LIRF actually uses fewer positive examples than the number of positive episodes, (110, 7K, and 1 actionable positive examples respectively) since the same example can be reused across episodes if it is not destroyed.
Finally, when feasible, we engineer an interactive reward function policy (Manual IRF) that performs hand-coded actions to evaluate the task: hand-coded unscrewing behavior for screwing, and a random poking action for the block stacking.\footnote{For Adroit hand door locking, the action space is too large to reasonably hand-code any IRF behavior.}
To isolate the effect of reward functions, all methods share the same input observation space for their policies.
See Appendix \ref{appendix:baselines} for more details on the baselines. 
Note that GAIfO, GT State Reward, and Manual IRF all involve more painstaking and comprehensive task specifications than LIRF, but offer useful comparison points nevertheless.

\subsection{Results}

\begin{table}[t]
    \centering
    \small
    \resizebox{0.9\textwidth}{!}{
    \begin{tabular}{ccLLL} 
     \toprule
       & Task Specification & Door Locking &  Block Stacking & Screwing  \\
     \midrule 
     VICE~\cite{fu2018variational} & Goal Examples & $0.032 \pm 0.015$ & $0.310 \pm 0.056$  & $0.02$  \\ 
     LIRF (Ours) & Actionable Goal Examples &  \textbf{0.640 $\pm$ 0.053} & \textbf{0.846 $\pm$ 0.027} & 0.82  \\ 
     LIRF+Verify (Ours) & Actionable Goal Examples &  \textbf{0.958 $\pm$ 0.007} & \textbf{0.884 $\pm$ 0.018} & \textbf{0.99}  \\
     \midrule
     GAIfO \cite{torabi2018generative} & Demonstrations & $0.112 \pm 0.024$ & $0.275 \pm 0.043$ & $0.0$  \\ 
     Manual IRF & Human Engineering & - & $0.613 \pm 0.058$ & $0.85$    \\ 
     GT State Reward & State+Human Engineering & $0.714 \pm 0.047$ & $0.956 \pm 0.021$ & $0.87$  \\ 
     \bottomrule 
    \end{tabular}
    }
    \caption{Task success rates of our method and baselines on the three tasks. 
    }
    \vspace{-5mm}
    \label{tab:baseline_table}
\end{table}

\paragraph{Does LIRF succeed in partially observable settings?} Table \ref{tab:baseline_table} shows task success rates for our approach and baselines.
LIRF outperforms VICE across all tasks, showing that LIRF successfully extends the exemplar reward framework into partially observable settings where prior exemplar rewards methods like VICE fail. LIRF also performs on par or better than baselines that use additional / more expensive forms of task supervision, such as demonstrations, human engineering, and privileged sensing of ground truth state.

\begin{figure}[t]
\centering
\includegraphics[width=\linewidth]{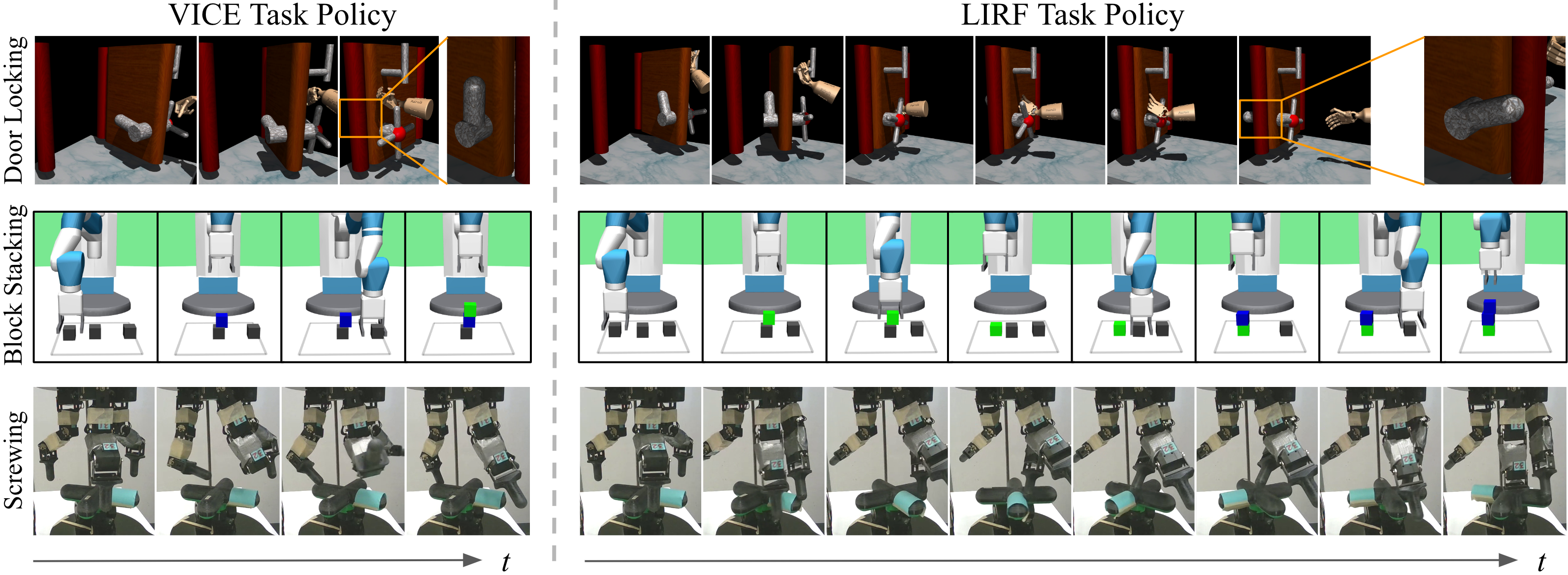}
\caption{LIRF task policy rollouts (right), compared to VICE (left). 
Note row 2, where VICE builds an unstable tower with the heavy block on top (visualized in green here, but this is not observable to the agent). LIRF self-corrects mid-episode to build the tower with heavy block on bottom.
See Supp slides for more video examples.
}
\vspace{-3mm}
\label{fig:pi_task}
\end{figure}

Qualitatively, LIRF policies show good task-solving behaviors for each task - they learn to lock the door, construct towers with the heaviest block at the bottom, and fully tighten the screw (see Figure \ref{fig:pi_task}). LIRF even demonstrates sophisticated self-correction behavior in the block stacking task by rebuilding the tower if the heavier block is incorrectly stacked on top of a lighter block. \jd{For fig 4, could you add the task names along the left, perhaps in vertical text for easier parsing?}

Among baselines, VICE and GAIfO perform poorly due to state aliasing, and LIRF is competitive or superior to Manual IRF, showing that \textit{learning} an IRF policy with RL may yield better rewards for training partially observable task policies.

\begin{wraptable}{r}{0.33\textwidth}
\vspace{-0.2in}
    \small
    \centering
    \begin{tabular}{cc} 
     \toprule
     Task & Accuracy \\
     \midrule 
     Door Locking &  $0.992 \pm 0.006$ \\
     Block Stacking & $0.852 \pm 0.023$ \\
     Screwing & $0.98$ \\
     \bottomrule
    \end{tabular}
    \caption{IRF task success classification accuracies.
    }
    \vspace{-3mm}
    \label{tab:verification_policy}
\end{wraptable}

\paragraph{Do the IRF policies $\pi_R$ sensibly evaluate task completion?}
First, we evaluate the success rate of the IRF policy $\pi_R$ at distinguishing between positive and negative examples for each task in Table~\ref{tab:verification_policy}. IRF-based task success classification is highly accurate for all tasks. Figure \ref{fig:pi_v} shows examples of learned reward function policies.  \jd{Could you color the thumbs-up and thumbs-down in green and red to make it visually more telling? Kun: hmm not sure how to do this}
The IRF policies learn sensible strategies to separate visually indistinguishable positives and negatives.
In the door locking task, IRF pulls the fixed handle; in stacking, it pokes the bottom block; in screwing, it applies a small counterclockwise torque to unscrew the valve.
Finally, IRF policy learning is quite efficient in terms of the number of actionable positive examples required. Door locking, screwing, and weighted block stacking need 110, 1, and 7k actionable positive examples respectively (note that the same positive example can be reused for multiple IRF training episodes, unless its ``positiveness'' is destroyed at the end of that episode). Block stacking is the hardest, since distinguishing stable from unstable towers requires a finely tuned poking force and motion, and even ``stable'' block tower examples are prone to toppling quite easily.
See Appendix \ref{appendix:irf_training} for more details. 

\begin{wrapfigure}[11]{r}{0.35\textwidth}
\vspace{-0.2in}
\centering
\includegraphics[width=0.35\textwidth]{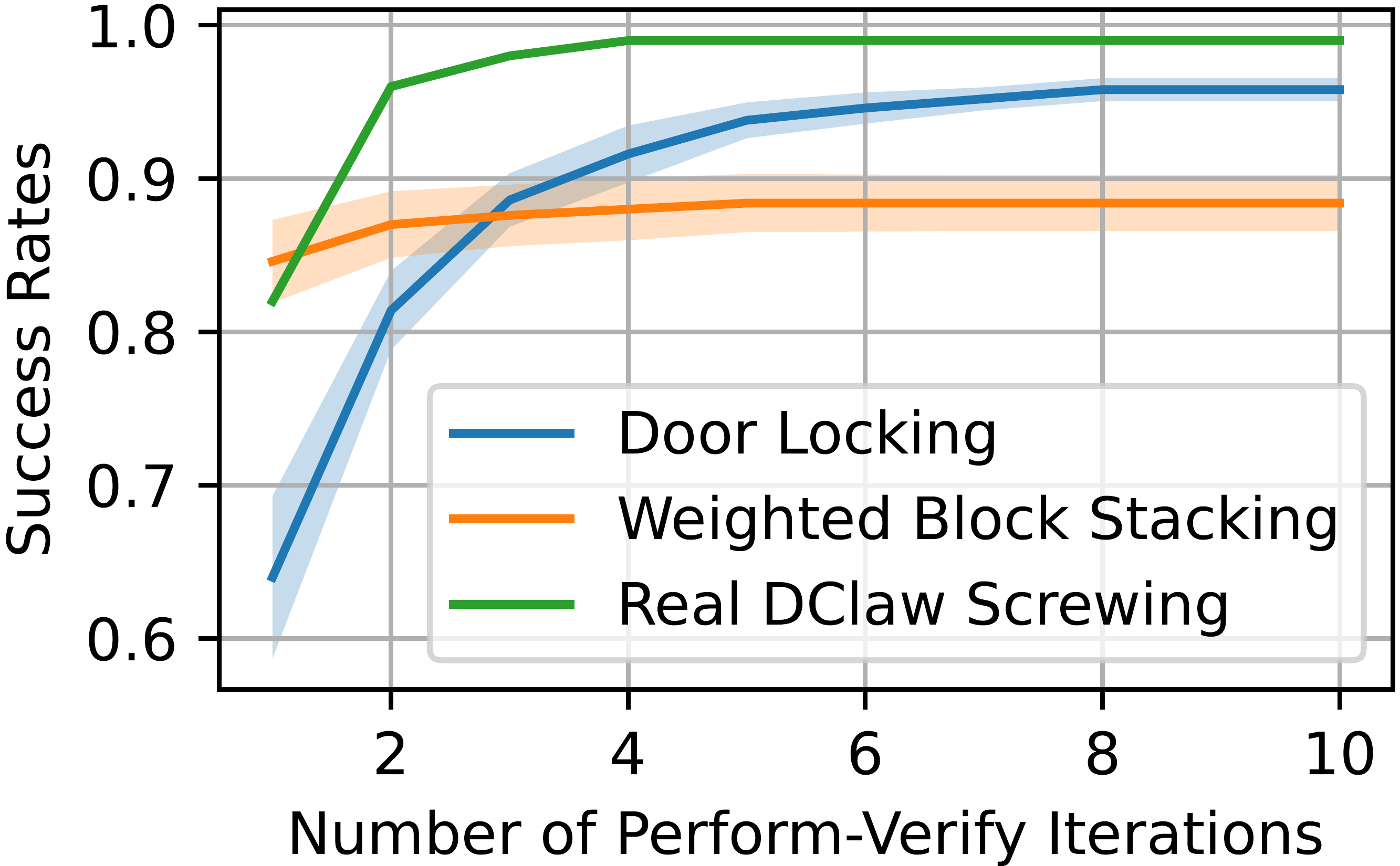}
\caption{LIRF+Verify success rates vs.~perform-verify iterations.
}
\label{fig:fsm_trend}
\end{wrapfigure}
\paragraph{Can the IRF policy be used for in-the-loop verification?}
As seen in Table \ref{tab:baseline_table}, LIRF+Verify, which reuses the learned IRF policy during task execution, outperforms all baselines, even beating the ground truth reward baseline in door locking and screwing. Figure~\ref{fig:fsm_trend} shows how LIRF+Verify improves performance with more perform-verify iterations.
LIRF+Verify makes the biggest difference in door locking: here, plain LIRF fails mainly because the policies operating on partial observations do not learn precisely how much to turn the latch without the aid of $\pi_R$ to declare task completion. This is also why GT State Reward performs quite poorly on this task despite training on the true state-based reward function (see also Appendix \ref{sec:gt-retry}). 
LIRF+Verify's gains over LIRF are smaller in block stacking and screwing, because LIRF failures here occur from low-level imprecision in pick-and-place or valve-turning. Here, LIRF+Verify helps by allowing the task policy $\pi_T$ to try again when it fails. We show video examples \edc{on the project website}. 

\begin{figure}[t]
\centering
\includegraphics[width=\linewidth]{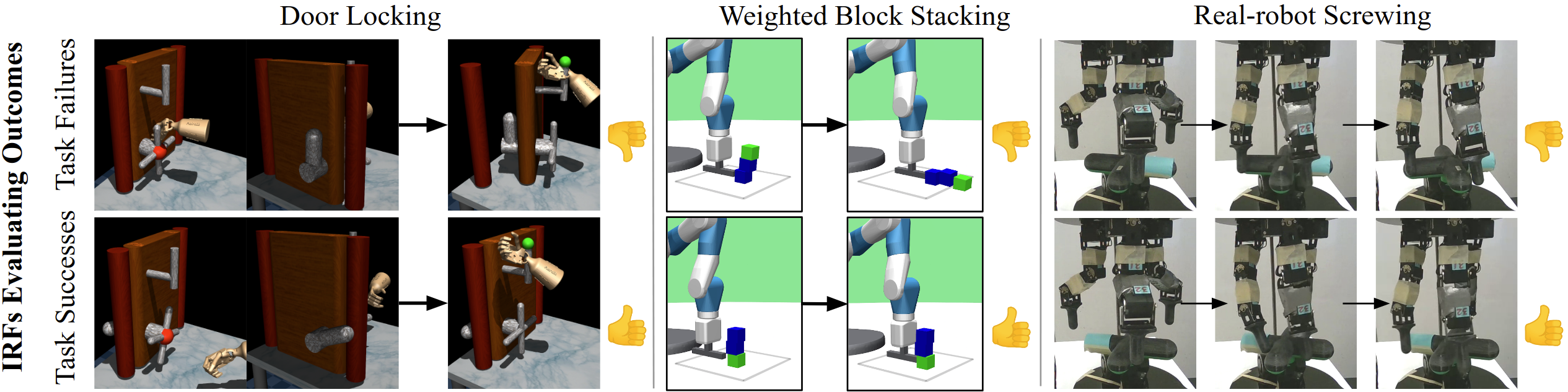}
\caption{Running the IRF policy $\pi_R$ on the success and failure actionable examples. Left (Door Locking): $\pi_R$ grasps the door and pulls to verify locking. Middle (Block Stacking): $\pi_R$ checks for tower stability by gently poking the bottom block. Right (Screwing): $\pi_R$ attempts to lightly unscrew - the screw in the top row rotates as it is not tight, whereas the bottom-row  does not.
}
\vspace{-5mm}

\label{fig:pi_v}
\end{figure}

\vspace{-3mm}
\section{Conclusions}
\vspace{-2mm}
We have presented LIRF, a framework for conveniently training example-based interactive robot policies to evaluate robot task policies, in order both to provide rewards to train them, and to verify their execution in partially observed tasks. While our results show substantial improvements over baselines, LIRF currently has two drawbacks. 
LIRF is best applied to settings where the initial task policies trained from single-image-based rewards fail most of the time. When this is not the case, such as in fully observable settings, IRF policies yield marginal or no benefits over the initial policies (Appendix \ref{sec:door-closing}), and therefore may not be the most frugal algorithmic choice.
Next, LIRF requires the physical storage of ``actionable outcomes'' as positive examples. While the number of such examples in some tasks may be small enough to not be a major concern (e.g., among our experiments, 1 for screwing, 110 for door locking), in other cases, storing and presenting a large number of positive example objects to train the IRF might be cumbersome (e.g., 7k for block stacking). We discuss scalability in more detail in Appendix \ref{sec:scalability}. 
Future work overcoming these limitations could further expand the domain of applicability of LIRF.

\clearpage

\section{Acknowledgement}
This work was supported by the U.S.~Office of Naval Research under grant ONR N00014-22-1-2677, and a gift from NEC Laboratories America.  

\bibliography{refs}  

\clearpage
\appendix
\appendix

\section{Appendix}
We encourage the reader to refer to \edc{our website at the url: \url{\projectwebsite}} for additional video visualizations.
\edc{Refer to \url{https://github.com/penn-pal-lab/lirf_public} for code release.}

\subsection{LIRF+Verify Approach Details}
\label{appendix:lirf_verify}

\edc{In Section 3.2, we proposed to use IRFs as verification mechanisms for introspective behaviors. Here we provide a detailed pseudocode for our LIRF+Verify approach.}

\begin{algorithm}[h]
\small
\caption{LIRF+Verify Pseudocode}
\begin{algorithmic}[1]
\renewcommand{\algorithmicrequire}{\textbf{Require:}}
\REQUIRE LIRF policy $\pi_T$, IRF policy $\pi_R$, single-observation reward $D(o) \rightarrow [0, 1]$
\STATE $i = 0$
\STATE Rollout $\pi_T$ in the task POMDP.
\STATE Keep the world state, switch to IRF POMDP and rollout $\pi_R$. The final observation is $o^*$.
\STATE $i{+}{+}$
\IF {$D(o^*) > 0.5$ or $i \geq 10$}
    \STATE Terminate
\ELSE
    \STATE Keep the world state, repeat from 1.
\ENDIF

\end{algorithmic}
\label{algo:main_algo_expanded}

\end{algorithm}

\subsection{Environment Details}
\label{appendix:environment}

\edc{In Section 5, we described the three tasks we use in our experiments. Here we provide supplementary minor details for each task. Refer to our code release for reproducibility.}

\textbf{Door Locking:} During initialization, the orientation of the latch $\theta$ is randomized from $\theta \sim (90\degree, 180\degree)$, with $0\degree$ pointing outwards from the door, $90\degree$ pointing upwards. The door is firmly locked when $\theta \in (-40 \degree, 40 \degree)$. The latch can freely rotate without joint limits, so that it is possible for a policy to overturn the latch.
An extra fixed handle is provided for the IRF policy so that it will not affect the state of the latch when trying to open to door.
The control of the Adroit hand outputs 5-D actions (x, y, z, roll, and close fingers).

\textbf{Weighted Block Stacking:} In this environment, three visually identical blocks are presented to the Fetch robot, with one block significantly heavier than the other two. The task phase is executed by a Fetch robot with a two-fingered gripper, and the IRF phase is executed by a Fetch robot with an end-effector more suitable for horizontal poking actions.

\textit{Motion primitives:} To expedite the RL training for the task policies and the IRF policies, we employ a ``pick-and-place'' motion primitive and a ``poking'' motion primitive respectively in the weighted block stacking task.

The task policies output a 4-dimensional action for each timestep, in which the first 2 dimensions define the $x$ and $y$ position in the world coordinate for where to pick up the block, and the last 2 dimensions correspond to the $x$ and $y$ position for where to place the block, assuming that a block is grasped. We predefine separate $z$ heights for the ``pick-up" actions and for the ``place'' actions. For visualizations of the ``pick-and-place'' primitive, refer to the supplemental videos.

\begin{figure}[h]
\centering
\includegraphics[width=0.48\textwidth]{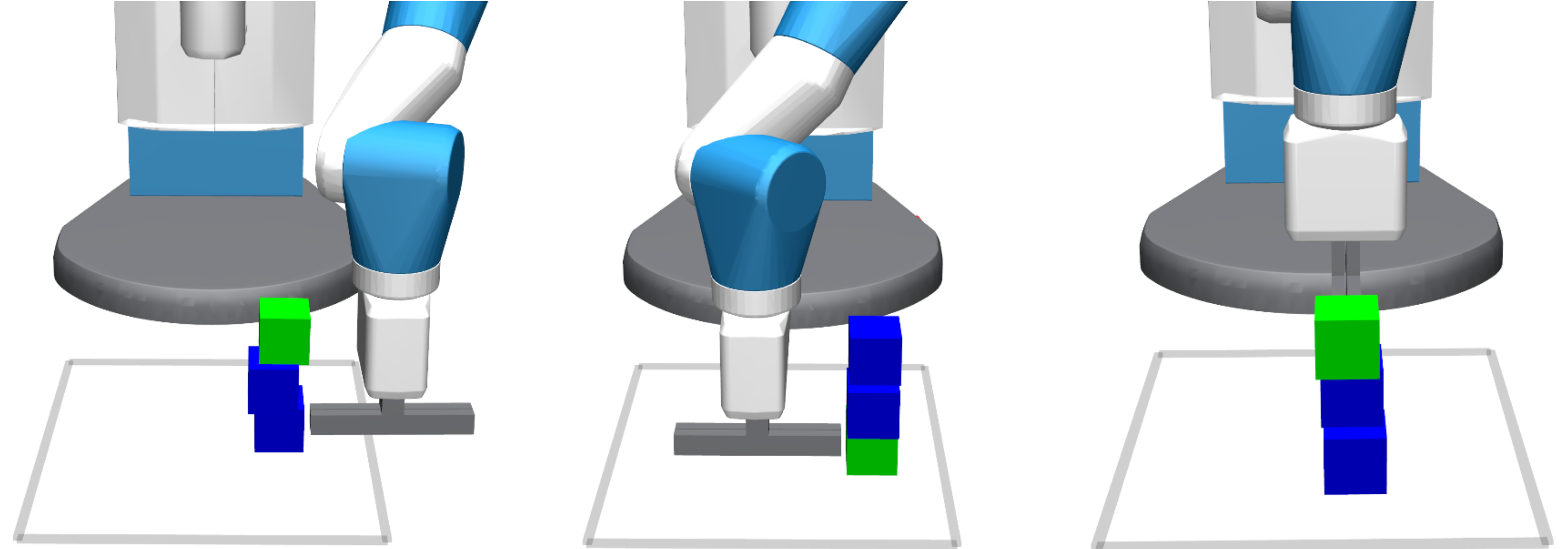}
\caption{Poking primitives for $\pi_R$.}
\label{fig:wbs_pi_R_primitive}
\end{figure}

To examine if a block tower is built in the correct order, i.e. the heavy block is put at the bottom, the robot pokes the tower to apply some disturbance and observes whether the block tower falls over. We modify the end-effector of Fetch to enable such poking actions as shown in Figure \ref{fig:wbs_pi_R_primitive}.
The IRF policies output a 3-dimensional action (poking height, direction, magnitude) for each timestep. See supplemental slides for more poking actions.

\textbf{Screwing: } The 4-prong valve is connected to a motor underneath it. The valve can rotate freely without limits when the motor is not engaged, whereas it is locked in place or controllable through the motor when the motor is engaged.

\subsection{Metrics}

\edc{In the main paper, we mainly use success rates to evaluate and compare all the policies. In this section, we describe how success rates are defined.}

Success rates for the task policies are measured based on ground truth state. Successful door locking corresponds to the latch orientation $\theta \in (-40\degree, 40 \degree)$, where it effectively locks the door. Successful block stacking corresponds to a standing tower of three blocks, with the heaviest block at the bottom. Successful screwing corresponds to the valve reaching the ``tightened" state.
When an IRF policy is executed starting from an initial partially observed state $\sigma$, its performance is reported based on its ability to classify $\sigma$. 
Concretely, for a state $\sigma$, we first execute $\pi_R$ to generate the new environment state $s^*$, which in turn generates a new observation $o^*$. Then we classify this observation using the single-observation reward classifier $D(o^*)$, and compare the result with the ground truth state-based success / failure label for $\sigma$.

\subsection{Baselines}
\label{appendix:baselines}

\edc{In this section, we provide additional details for the baselines described in Section 5.
}

\subsubsection{Manual IRF Baseline Details}
To demonstrate that the IRF policies are non-trivial to learn, we also compare our method against task policies trained against manual IRF policies, in which we manually engineer IRF policies to distinguish between goal states and failure states.
Since it is difficult to engineer a motion primitive for the Adroit hand to open the door, we only evaluate the ``Manual $\pi_R$" baseline on the weighted blocking task and the real-robot screwing task. 

\paragraph{Screwing} Here, we program a simple hard-coded movement of the hand to perturb the valve. 

\begin{figure}[h]
    \centering
    \includegraphics[width=0.35\textwidth]{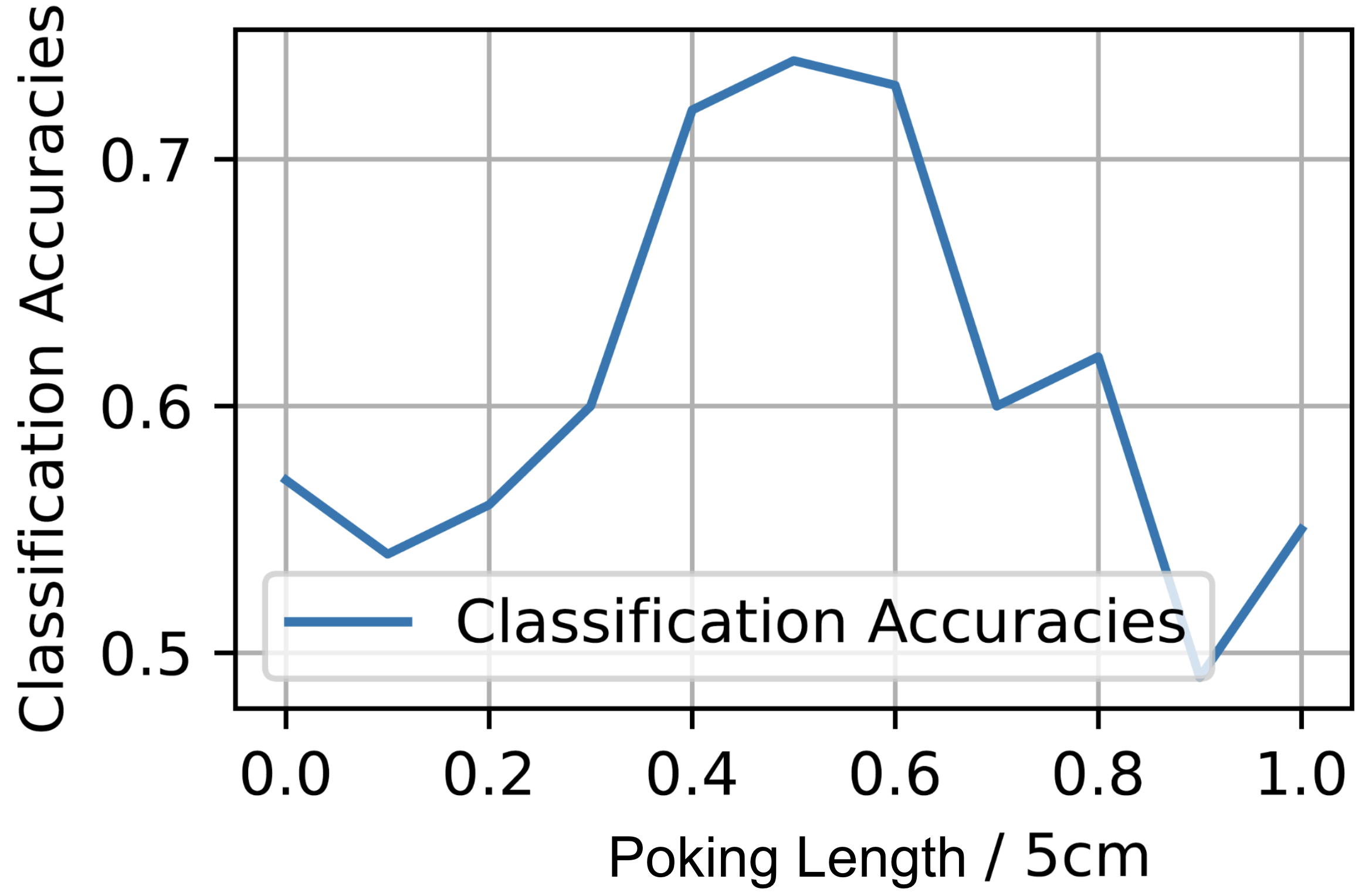}
    \caption{The performance of random $\pi_R$ vs. $\pi_R$ poking magnitude in weighted block stacking.}
    \label{fig:wbs_engin_pi_R}
\end{figure}

\paragraph{Weighted Block Stacking} For this setting, we grid search for the best poking length, i.e.  the one that best separates stable and unstable stacks, while using random poking heights and poking directions. 
The random $\pi_R$ classification accuracy vs. poking length is shown in \ref{fig:wbs_engin_pi_R}. We choose the poking length with the highest classification accuracy ($2.5$ cm) as our manual $\pi_R$.

\subsubsection{GAIfO Baseline Details}
Where do expert demonstrations come from for the GAIfO imitation approach \cite{torabi2018generative} used in our comparisons? For both door locking and real-robot screwing, we collect observation-only expert trajectories as demonstrations from policies trained on GT-state observation (without aliasing) against GT reward. 

For weighted block stacking, we hand-code an expert that always places the heavier block at the bottom, and places the other two blocks on top of the heavier block to build a block tower. The observations do not contain the block weights, to mimic a purely visual observation.
However, for thoroughness, we also perform additional experiments when demonstration observations include block weight information for blocks that have been picked up within the demonstration. Naturally, GAIfO benefits from this, and its success rate improves from $0.275$ to $0.780$, still lower than our LIRF approaches.

\subsubsection{GT State Reward Baseline Details} 
For all three tasks, the GT State Reward Baseline policies \jd{Is this not easy to squeeze into the main paper? Kun: I think in the main paper we implicitly pointed out that all task policies share the same obs, so here it's just re-stating.} 
are trained with the same observation as LIRF polices against sparse rewards given at success.

\subsection{LIRF Task Policy Training: Implementation Details}
\label{appendix:lirf_training}

\edc{Here we provide additional implementation details for the LIRF task policy described in Section 5.}

We use fully-connected MLPs with three hidden layers of 128 units and ReLU nonlinearities for all policies, critics and reward functions in the door locking experiment and real-robot screwing experiment.

\paragraph{Weighted Block Stacking} 
To mimic visual observations of the identical blocks, where block identities would not be visible before interaction, we employ a permutation-invariant network (Deep Sets \cite{zaheer2017deep}) for both policies and VICE success classifiers, so that
\begin{equation}
    \pi([\vec{x}_0, \vec{x}_1, \vec{x}_2]) = \pi([\vec{x}_{perm(0)}, \vec{x}_{perm(1)}, \vec{x}_{perm(2)}])
\end{equation}
where $\vec{x}_i$ represents the 3D position and weight (if known) of block $i$, $perm(i)$ is a permutation function of $i$. \jd{I don't understand what the equation is supposed to convey. Kun: this is how Deep sets define permutation invariant} Our neural networks for all policies, critics and reward functions contain the followings: two hidden layers of 128 units and ReLU nonlinearities, then one AvgPool1d along the object dimension, finally two layers of 128 units and ReLU nonlinearities.
The weight for each block is initialized as $-1$. When a block is picked up, the weight for it becomes $1$ if it is the heavier block, $0$ otherwise.

\subsection{Ablation Study on $\lambda$}\label{sec:lambda} 
We also carry out ablation studies on the hyper-parameter $\lambda$, which is the weight for the IRF reward, in the door locking setting. The results are shown in Table \ref{tab:lambda-table}. As the IRF reward is only provided sparsely for the terminal state of the task policy, and the original single-observation reward $D(o_t)$ is provided for each step in the episode, we need $\lambda$ to be large enough in order to have a substantial effect during LIRF training. Beyond that, the performance of LIRF is not very sensitive to $\lambda$.

\begin{table}[h]
    \centering
    \small
    \begin{tabular}{cc} 
     \toprule
     LIRF Policy &  Door Locking Success Rate \\
     \midrule 
     $\lambda = 10$ & $0.32$  \\ 
     $\lambda = 100$ & $0.54$  \\ 
     $\lambda = 1000$ & $0.64$  \\ 
     $\lambda = 10000$ & $0.62$  \\ 
     \bottomrule 
    \end{tabular}
    \caption{Door locking success rates of LIRF policies trained with different $\lambda$. 
    }
    \label{tab:lambda-table}
\end{table}

\subsection{IRF Policy Training Efficiency}
\label{appendix:irf_training}
\edc{In Section 5, we discussed the number of positive examples that is required for IRF policy training. Here we provide more details for IRF policy training efficiency.}

\begin{figure}[h]
    \vspace{-0.2in}
    \centering
    \includegraphics[width=0.35\textwidth]{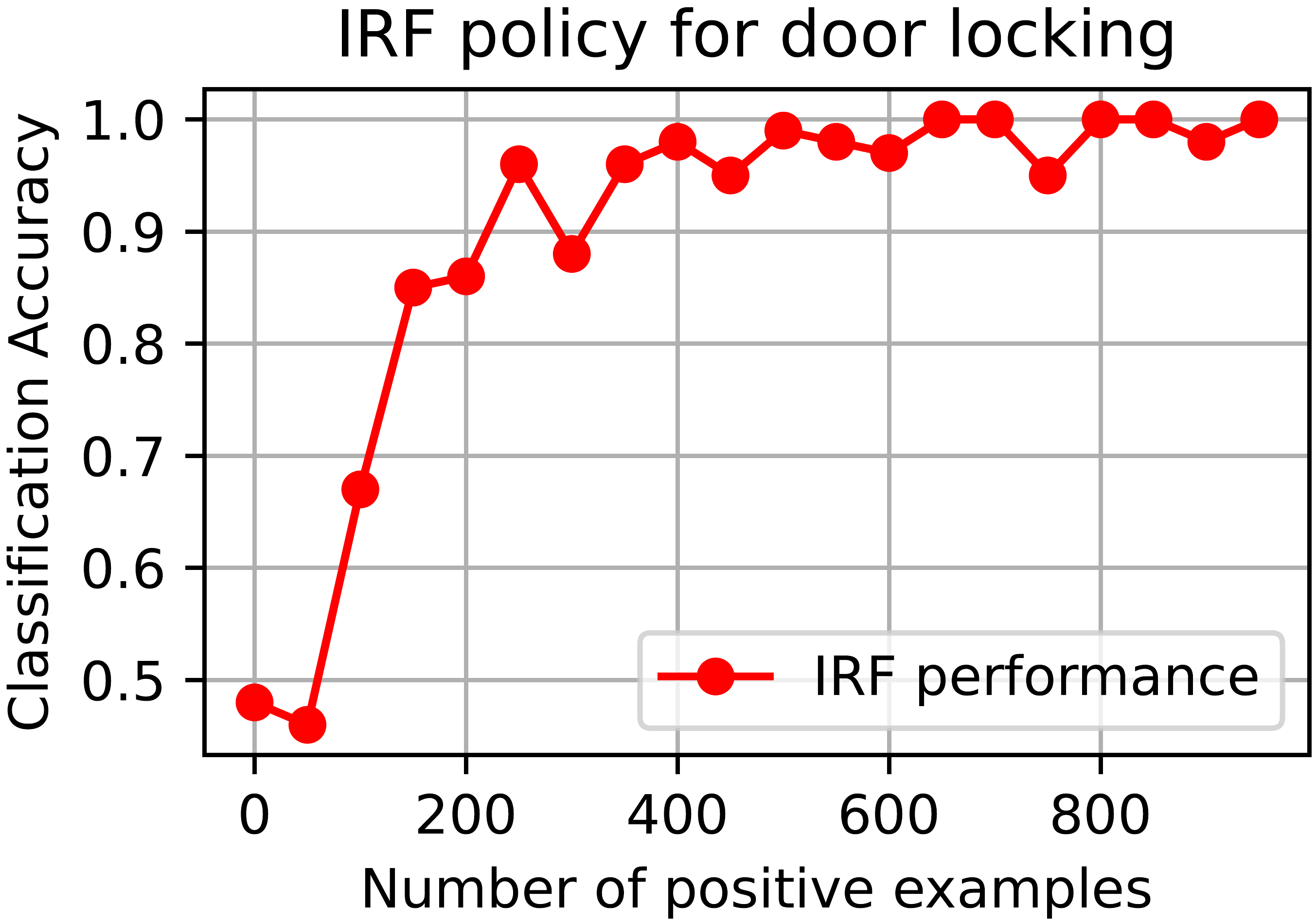}
    \caption{$\pi_R$ classification accuracy vs. number of positive examples given in the door locking task.}
    \vspace{-3mm}
    \label{fig:pi_R_pos_examples}
\end{figure}

To evaluate the efficiency of our example-based interactive reward function training, we record the number of positive examples that are required for training an accurate IRF policy. Figure \ref{fig:pi_R_pos_examples} shows the training curve of $\pi_R$ classification accuracy vs. number of positive examples given. The IRF policy is able to reach higher than $90\%$ classification accuracy with fewer than $300$ positive examples in the door locking task.

The performance of IRF converges after 250, 10k and 100 episodes with positive examples for door locking, weighted block stacking, and screwing tasks respectively. Note that since actionable positive examples are reusable across IRF training episodes unless they are destroyed, we observe that these three tasks use 110, 7k, and 1 actionable positive example respectively. 
Finally, we use a balanced sampling strategy, so the number of episodes with negative examples is roughly equal to the number of episodes with positive examples. Negative actionable examples for training the IRF policy are simply the outcomes of the current task policy, generated on-the-fly for each training episode.

\jd{Can we show some training curves for the task policy somewhere, over different phases, with perhaps some demarcations for the initial VICE phase, and the final LIRF phase? Ideally, it would be good to see rewards rising and stagnating for VICE, then climbing higher once LIRF starts.}

\subsection{Can the GT state reward baseline also benefit from multiple retries?}\label{sec:gt-retry}

We showed in Section 5 Table \ref{tab:baseline_table} that LIRF verification policies can be used not just to train task policies but also to improve test-time task performance through identifying when the task policy has failed and should retry. The ground truth (GT) state reward baseline in the main paper doesn't directly permit retries since rewards in RL are typically assumed available only at training time, and the task policy for this baseline is assumed to only have access to observation histories. To understand how retries affect the performance of the GT state reward baseline policies, we evaluate two alternative approaches:
\begin{enumerate}[leftmargin=*]
    \item First, we run the GT state reward baseline with increased episode length so that its total duration matches the K retries the LIRF policy gets.
    \item Second, we provide ground truth state reward access to check for task success instead of our IRF reward policy. This GT state reward policy with GT state reward checking would correspond to an ``oracle" version of our method.
\end{enumerate}

The results for GT state reward baselines with multiple retries are shown in Table \ref{tab:gt-retry-table}. These experimental results align with our expectation that, without GT state success checking, the performance of GT state reward baselines does not improve with more retries since they tend to overturn the door latch. With GT state success checking, the GT state reward policies with multiple retries stop right after they successfully lock the door, achieving better performance, as they serve as an ``oracle" version of our method. 

\jd{Also include our method performance from the main paper into this table, for easy comparison.}

\begin{table}[h]
    \centering
    \small
    \begin{tabular}{cc} 
     \toprule
     Policy &  Door Locking Success Rate \\
     \midrule 
     GT state reward without retries & $0.71$ \\
     GT state reward with 3 retries & $0.71$  \\ 
     GT state reward with 10 retries & $0.60$  \\ 
     GT state reward with 3 retries and GT state success checking & $0.91$  \\ 
     GT state reward with 10 retries and GT state success checking & $0.98$  \\ 
     \midrule 
     LIRF (Ours) & $0.64$ \\
     LIRF+Verify (Ours) & $0.96$ \\
     \bottomrule 
    \end{tabular}
    \caption{Door locking success rates of GT state reward baselines with multiple retries. 
    }
    \label{tab:gt-retry-table}
\end{table}

\subsection{LIRF When State is Fully Observed: Door Closing Task}\label{sec:door-closing}
In the main paper, we evaluated LIRF in partially observed state-aliased settings where naive baselines like VICE fail. For example, for our door locking task, the visual appearance of the locked door is indistinguishable from that of a closed but unlocked door. As such, VICE would learn policies that merely close the door. 

To evaluate LIRF in a fully observed setting where VICE already achieves near-perfect performance, we select the door \textit{closing} task. From Table \ref{tab:door-closing-table}, We can see that the VICE policy achieves near-perfect performance on the door closing task as it is fully-observed and relatively easy to learn, but the performance of LIRF does not degrade under such circumstances. \jd{Success rates measured over how many trials? Is performance of VICE really 100\%, and performance of LIRF really worse? Or is this within margin? Just looks a bit weird the way it is now. Kun: 100 trials each. I think here the main point is that for this easy to learn task, both VICE and LIRF are able to solve the task consistently.}

\begin{table}[h]
    \centering
    \small
    \begin{tabular}{cc} 
     \toprule
    Policy &  Door Closing Success Rate \\
     \midrule 
     VICE & $1.0$  \\ 
     LIRF & $0.99$  \\ 
     \bottomrule 
    \end{tabular}
    \caption{Door closing success rates of VICE and LIRF. 
    }
    \label{tab:door-closing-table}
\end{table}

\subsection{Positive Examples and Scalability}\label{sec:scalability}
In the limitations section, we pointed out that there are cases in which the number of actionable positive examples required by LIRF is large, and gathering and storing them is non-trivial. We now discuss scalability in more detail.

\textbf{First, what causes the number of required actionable positive examples to be high or low?} In the weighted block stacking task, the desired task outcome, a well-constructed stack of blocks, is not objectively a very stable object configuration (even if it is more stable than poorly constructed stacks) — in particular, relatively small perturbations can destroy it. As a result, we use around 7k positive examples, over 10k IRF training episodes (the remaining 3k episodes don’t result in the destruction of a positive example so that we can reuse the examples provided). Further, it takes a long time for the interactive reward function (IRF) to learn the nuanced poking behavior that can accurately separate ``good" block stacks from ``bad", to then produce a good task policy. The IRF policy must apply enough force to the tower at the right locations to reveal the weights of the blocks, but be gentle enough to avoid toppling the tower. At the other extreme, in the screwing task, we required only one single positive example of a tightened screw, over 100 IRF training episodes with positive samples: here, the perturbation behavior to separate loose and tight screws is very straight-forward to learn, and the positive examples are not easily destroyed.

\textbf{Second, how should we think about the cost of training LIRF policies in these terms?} When the IRF policy does not destroy a positive example, it can be reused. For example, in the screwing task, technically only one single positive example of a tightened screw is required because the positive example physically cannot be destroyed (undoing a tight screw) by the robot during training, so that this one tightened screw can be reused for 100 times. Note that in the paper, we reported 250, 10k and 100 trials with positive samples for training IRF policies for door locking, weighted block stacking, and screwing tasks respectively. However, if we take reusability of positive examples into consideration, the IRF training only requires 110, 7k, and 1 positive examples.

\textbf{Finally, how could LIRF be encouraged to learn from fewer positive examples?} Here, we speculate that explicitly penalizing the destruction of positive examples and/or explicitly encouraging the verification policy to generate reversible behaviors will help. Further, we have used off-the-shelf policy learning approaches (SAC) without optimizing for sample efficiency, and there may be significant gains from using alternative, more sample-efficient approaches, such as from the model-based reinforcement learning literature.

\end{document}